\documentclass[letterpaper, 10 pt, conference]{ieeeconf}  % Comment this line out if you need a4paper

\IEEEoverridecommandlockouts                              % This command is only needed if 
\usepackage{amsmath,amssymb,amsfonts}
\usepackage{graphicx}
\usepackage{multirow}
\usepackage{booktabs}

\title{\LARGE \bf
DynaVIG: Monocular Vision/INS/GNSS Integrated Navigation and Object Tracking for AGV in Dynamic Scenes
}

\author{Ronghe Jin$^{1}$, Yan Wang$^1$, Zhi Gao$^2$, Xiaoji Niu$^1$, Li-Ta Hsu$^3$, and Jingnan Liu$^1$ % <-this % stops a space
\thanks{This work was supported by the National Key Research 
and Development Program of China under Grant 2016YFB0501804.
(Corresponding author: Ronghe Jin.)}
\thanks{$^{1}$Ronghe Jin, Yan Wang, Xiaoji Niu, and Jingnan Liu are with GNSS Research Center,
        Wuhan University, No. 129 Luoyu Road, Wuhan 430079, China
        {\tt\small \{huanhexiao,wystephen,xjniu,jnliu\}@whu.edu.cn}}%
\thanks{$^{2}$Zhi Gao is with the School of Remote Sensing and Information Engineering, Wuhan University, No. 129 Luoyu Road, Wuhan 430079, China
        {\tt\small gaozhinus@gmail.com}}%
\thanks{$^3$Li-Ta Hsu is with the Department of Aeronautical and Aviation Engineering,
The Hong Kong Polytechnic University, Hong Kong SAR, China
{\tt\small lt.hsu@polyu.edu.hk}}  
}

\begin{document}

\maketitle
\thispagestyle{empty}
\pagestyle{empty}

%%%%%%%%%%%%%%%%%%%%%%%%%%%%%%%%%%%%%%%%%%%%%%%%%%%%%%%%%%%%%%%%%%%%%%%%%%%%%%%%
\begin{abstract}

        Visual-Inertial Odometry (VIO) usually suffers from drifting over long-time runs, 
        the accuracy is easily affected by dynamic objects.
        We propose DynaVIG, a navigation and object tracking system based on the integration of Monocular Vision, Inertial Navigation System (INS), and Global Navigation Satellite System (GNSS).
        Our system aims to provide an accurate global estimation of the navigation states and object poses for the automated ground vehicle (AGV) in dynamic scenes.
        Due to the scale ambiguity of the object, 
        a prior height model is proposed to initialize the object pose, 
        and the scale is continuously estimated with the aid of GNSS and INS. 
        To precisely track the object with complex moving, 
        we establish an accurate dynamics model according to its motion state. 
        Then the multi-sensor observations are optimized in a unified framework.
        Experiments on the KITTI dataset demonstrate that the multi-sensor fusion can effectively improve the accuracy of navigation and object tracking,
        compared to state-of-the-art methods. 
        In addition, 
        the proposed system achieves good estimation of the objects that change speed or direction.

\end{abstract}

%%%%%%%%%%%%%%%%%%%%%%%%%%%%%%%%%%%%%%%%%%%%%%%%%%%%%%%%%%%%%%%%%%%%%%%%%%%%%%%%
\section{INTRODUCTION}

Navigation and object tracking are two significant tasks in autonomous driving and robotics.
Simultaneous Localization and Mapping (SLAM) using a monocular camera has low cost and high computational efficiency.
The fusion of monocular SLAM with Inertial Navigation System (INS) and Global Navigation Satellite System (GNSS) can greatly improve the accuracy and robustness of navigation,
the scale estimation enables monocular SLAM to obtain the capability of 3D measuring, 
which is similar to stereo or LiDAR.
Object tracking can obtain the object's pose, 
allowing safety in automatic driving and physical interaction in augmented reality (AR)/virtual reality (VR).
SLAM and object tracking are strongly correlated,  
some studies \cite{VDO-SLAM-2021}-\cite{TwistSLAM-2022} 
have recently unified the problem of SLAM and object tracking and verified that they can benefit each other.

Many researchers studied Visual-Inertial Odometry (VIO) for the complementarity of the Inertial Measurement Unit (IMU) and SLAM.
However, VIO has four unobservable directions \cite{caoGVINS-2021} and suffers from drifting over long-time runs. 
GNSS is an easy-obtained, drift-free, and global-aware observation that provides accurate long-term correction, 
thus Vision/INS/GNSS integration becomes attractive.
The integration mainly includes loosely-coupled integration using GNSS position results and tightly-coupled integration using GNSS raw measurements \cite{tangICGVINS-2022}, 
they achieve similar accuracy in open environments.
Loosely-coupled integration is convenient to design the algorithm and configure the information matrix, 
while GNSS cannot provide results with less than 4 satellites.
Tightly-coupled integration can work in challenging scenes using even 1 satellite, 
but the insufficient observations and multipath effect will seriously reduce the accuracy \cite{wenTCGI-2019}. 
Moreover, the framework of tightly-coupled integration is complex, and the noise propagation needs careful handling.
A general problem is that most VIO and Vision/INS/GNSS integrated systems neglect dynamic objects, 
which will reduce the performance in dynamic scenes.

\begin{figure}[!t]
	\centering
	\includegraphics[width=0.48\textwidth]{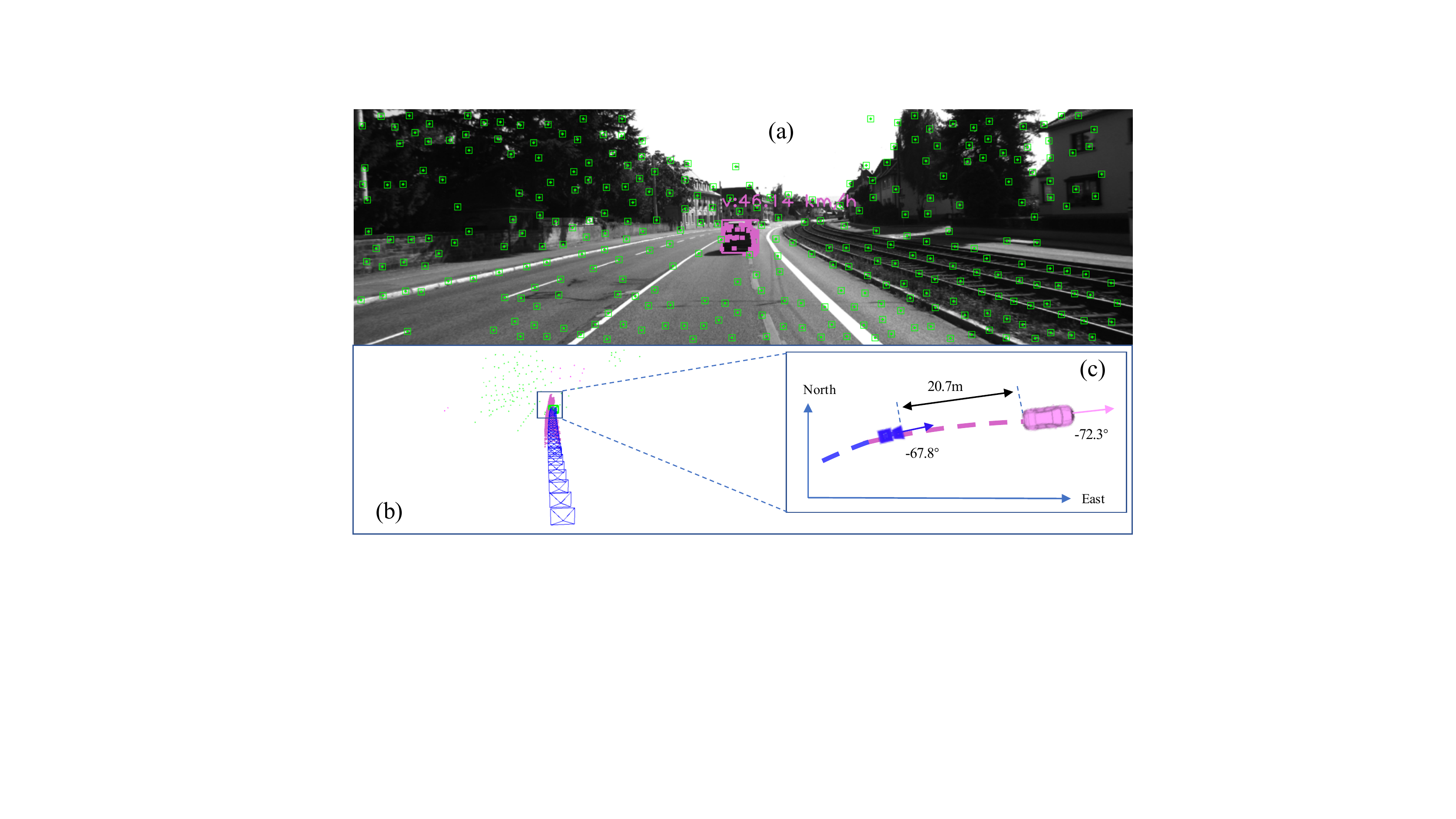}
	\caption{One example on KITTI shows: 
	(a) one object (pink) with its speed, and the static features (green).
	(b) 4D map corresponding to (a), trajectories of the camera (blue) and the object(pink).
	(c) The projection of the camera and the object on a 2D plane, the yaw angles and the object depth are given. }
	\label{Abstract}
\end{figure}

To decrease the influence of dynamic objects, 
some works detect and eliminate them. 
However, the simple elimination may lose some available information about the objects. 
Some recent works \cite{VDO-SLAM-2021}-\cite{TwistSLAM-2022}, \cite{CubeSLAM-2019} have unified SLAM and object tracking and achieved a win-win for such two tasks, 
however, there are some shortcomings. 
The works with the monocular camera usually use the camera height to scale the map and object, 
but this needs a changeless camera height and an observable ground plane. 
The object pose of the 6 Degree of Freedom (DoF) definition fails to exploit the constraints of plane ground, 
while the 3 DoF definition ignores slopes. 
In addition, most works assume a constant velocity model of the camera and objects, 
which will affect the accuracy when they change speed or direction. 

To address the above issues,
we propose DynaVIG based on the Monocular Vision/INS/GNSS integration for navigation and object tracking. 
A loosely coupled GNSS is applied considering the complexity of object tracking and the availability of the KITTI dataset. 
The object pose is defined as 4 DoF to make better use of the ground constraint, 
and it is initialized via a prior height model due to the scale ambiguity.
An accurate dynamics model of the object is constructed to process objects with complex motion, 
it is then combined with multi-sensor measurements to optimize the navigation states, map points, and object poses. 
Experiments of the KITTI tracking dataset are conducted for validation, 
one example is shown in Fig. \ref{Abstract}.
We highlight the contributions of our work as follows: 
\begin{itemize}
	\item A unified framework of navigation and object tracking is constructed based on the Monocular Vision/INS/GNSS integration; 
	\item A prior height model and a precise dynamics model of the object are proposed for accurate object tracking;
	\item The experiments verify the improvements of the proposed system compared with existing methods;
\end{itemize}

\section{RELATED WORK}

\begin{figure*}[!t]
	\centering
	\includegraphics[width=0.99\textwidth]{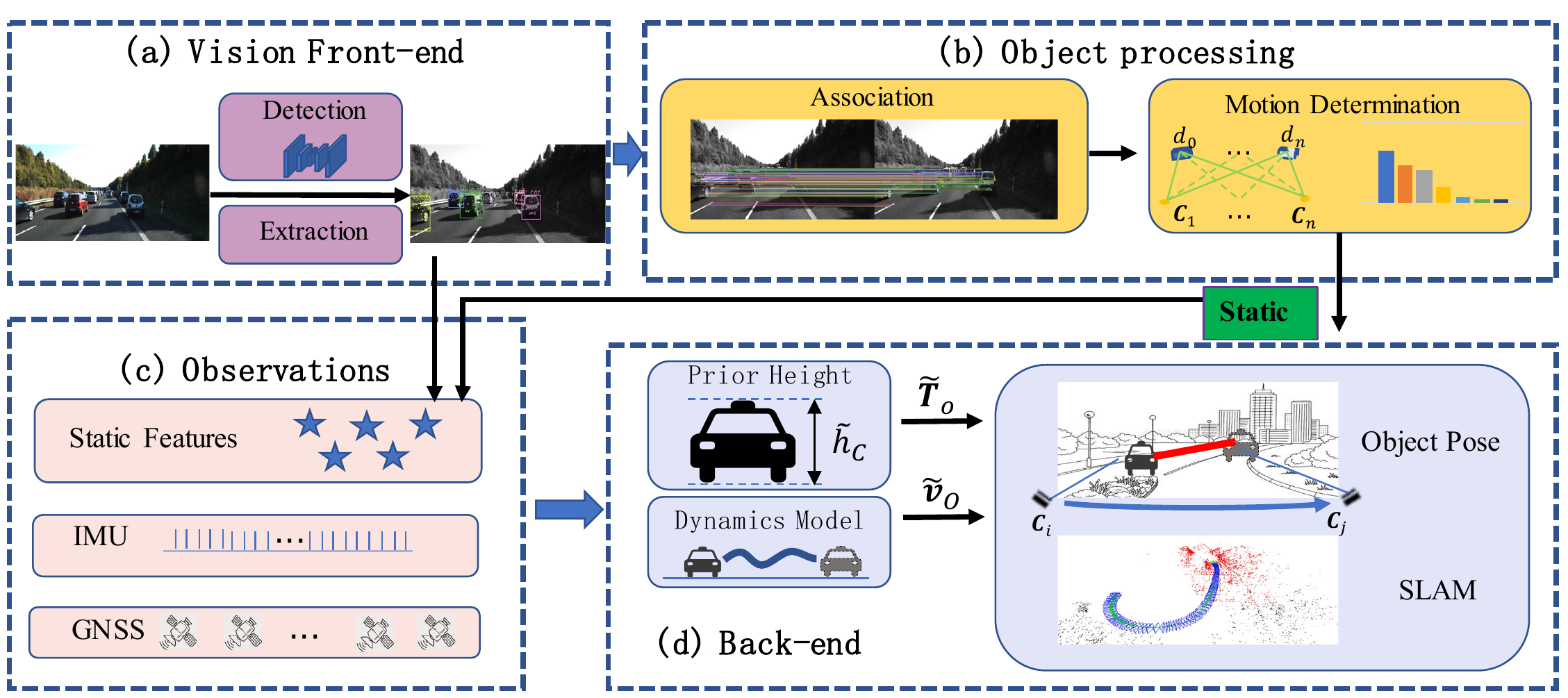}
	\caption{Overview of our proposal. 
		 The front-end generates the objects with YOLOv5 and extracts Good Features, as shown in (a).
		(b) demonstrates the object association via optical flow with the BRIEF descriptor, 
		its motion state is determined by the statistical characteristics of its depth sequence. 
		(c) illustrates the observations including static features extracted in (a) and static objects in (b), as well as IMU and GNSS. 
		(d) shows that given the prior height and dynamics models of the object, SLAM and object tracking are jointly optimized.}
	\label{DynaVIG}
\end{figure*}

\subsection{Visual SLAM with Multi-Sensor Fusion}

VIO is a widely researched topic \cite{VINSMono-2017}-\cite{ORBSLAM3-2021} and obtains great improvements,
but it suffers from drifts and unobservable directions. 
Scholars have studied the Vision/INS/GNSS integration to overcome the weaknesses of VIO. 
Earlier works are mainly loosely-coupled integrations.
VINS-Fusion \cite{qinVINSFuision-2019} couples GPS positions with VIO poses, 
but the result-level fusion depends heavily on the quality of GPS and VIO outputs. 
The work in \cite{CioffiVIG-2020} uses GNSS to couple with INS and vision, 
however, the GNSS simulated from the indoor dataset may limit the application. 
Recent works researched tightly-coupled integration to make better use of GNSS raw measurements.
GVINS \cite{caoGVINS-2021} is an excellent work of GNSS tightly-coupled integration with VIO, 
but it uses low-precision GNSS pseudorange measurements with meters of noise, 
and the ionospheric delay and troposphere delay using standard models may not be accurate enough.
GAINS \cite{LeeGAINS-2022} uses GNSS pseudorange, Doppler frequency shift, and carrier phase measurements with a lightweight filter, 
which could be prone to the nonlinear error of SLAM. 
The main advantage of tightly-coupled integration is the ability to provide continuous service in challenging scenes, 
but the model is complicated and the accuracy may still be limited.

\subsection{Scale Estimation for Monocular SLAM}

Scale estimation is a critical topic for monocular SLAM. 
The work in \cite{BayesianScale-2018} estimates the scale of monocular SLAM with a Bayesian filter, 
it uses the camera height to provide the initial scale and the object's prior height for correction. 
CubeSLAM \cite{CubeSLAM-2019} also uses the camera height and object size for scale. 
But different from \cite{BayesianScale-2018}, 
CubeSLAM constructs a framework for SLAM and objects to maintain a consistent scale.
These approaches assume a given fixed camera height and an observable ground plane, 
which is easily influenced by shaking, occlusion, and slope.
Therefore, some scholars try to recover the monocular scale without prior information.
The work in \cite{ScaleSyntheticData-2019} uses some network architectures to estimate absolute distances between consecutive frames. 
The authors of \cite{ObjectSupplementedScale-2018} introduce the concept of "extent" to constrain the scale drift of SLAM and objects.
These methods do not need constant camera height or planar roadway, 
but there is no information for physical scale estimation.

\subsection{Dynamic SLAM with Object Tracking}
To weaken the impact on SLAM, 
earlier works use geometric \cite{ORBSLAM-2015,ORBSLAM2-2017} 
or learning-based methods \cite{DynaSLAM-2018,xiaoDynamic-SLAM-2019}
to remove the dynamic objects.
These works are effective but lose high-level information, 
failing to maximize the SLAM accuracy. 
Recently, researchers make efforts to couple the problems of SLAM and object tracking.
CubeSLAM \cite{CubeSLAM-2019} generates the object's 3D bounding box using the 2D bounding box and vanishing points, 
then SLAM and object tracking are optimized together. 
CubeSLAM realizes 3D object detection with only one camera, 
but it is limited to stationary or slow-moving objects.
VDO-SLAM \cite{VDO-SLAM-2021} uses dense optical flow to ensure the robustness of object tracking, 
however, the calculation is very complicated.
DynaSLAM II \cite{DynaSLAMII-2021} proposes a tightly-coupled algorithm of SLAM and object tracking, 
but the object pose is defined as 6DoF without the constraint of the ground.
TwistSLAM \cite{TwistSLAM-2022} uses plane ground assumption to constrain an object's movements, 
the performance shows great advantages over previous works, 
while the 3DoF of pose definition may not satisfy the slopes. 
Moreover, These works use a constant velocity model of the camera and object, 
which may be inaccurate in some cases.
Some algorithms treat all objects as dynamic, 
resulting in fewer available features when the object is static. 

Most multi-sensor integrated approaches are easily affected by dynamic objects, 
and the accuracy of SLAM and object tracking algorithms are usually limited, 
thus the navigation performance of the automated ground vehicle (AGV) could be seriously restricted in dynamic scenes.
To this end, 
we aim to build an accurate global navigation and object tracking system using the Monocular Vision/INS/GNSS integration. 
By leveraging the drift-free GNSS and high-rate INS measurements, 
the system can eliminate the drift of SLAM and enable 3D object tracking with a monocular camera. 
The system can be used for AGV to precisely estimate the poses of the camera and objects.

\section{METHOD}

The structure of the proposed system is shown in Fig. \ref{DynaVIG}.
After being detected by YOLOv5, 
the objects are associated between frames by optical flow with the BRIEF descriptor. 
The object's motion state is rapidly determined, 
and the static ones are regarded as a part of the environment. 
IMU pre-integration and GNSS solutions can be calculated with parallel threads.  
Before the optimization,  
the prior height model is used to initialize the object pose, 
and the dynamics model is established according to its motion state. 
Then the multi-sensor measurements are optimized for the navigation and object tracking in a unified framework.

\subsection{Notations}\label{NotationSect}

\begin{figure}[!t]
	\centerline{\includegraphics[width=0.49\textwidth]{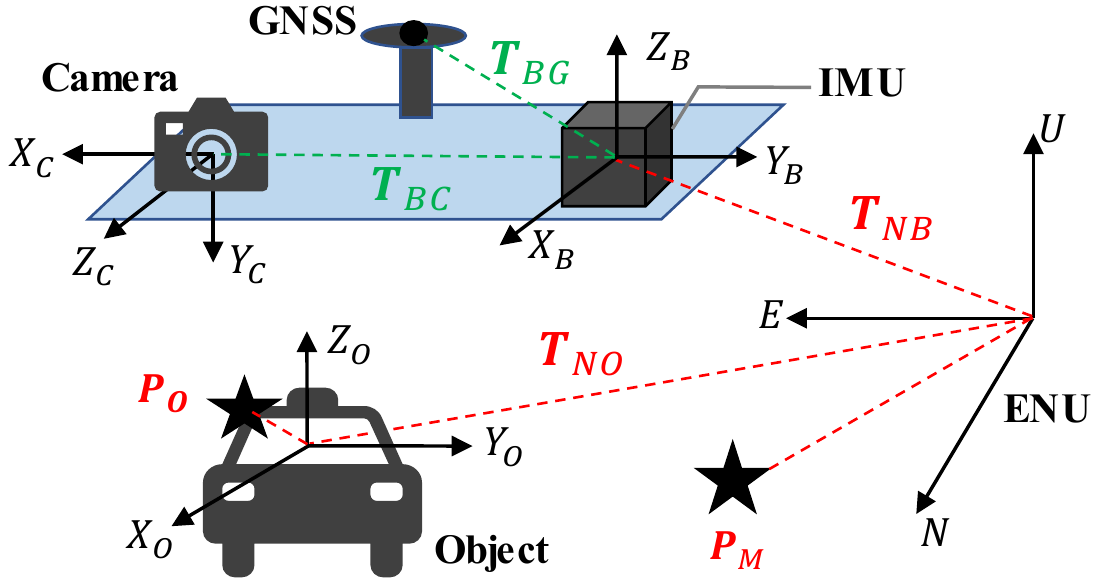}}
	\caption{Illumination of the coordinate frames, 
	including the frame $B$, $C$, $O$, and $N$ (ENU).
	The green symbols are the known extrinsic parameters between sensors, 
	and the red symbols are the states. }
	\label{Notation}
\end{figure}

We define $\boldsymbol{T}_{XY}\in SE(3)$ as the transformation from frame $Y$ to $X$,
$\boldsymbol{P}_{X}\in \mathbb{R}^3$ as the point coordinate in frame $X$,
and $\boldsymbol{v}_{XY}\in \mathbb{R}^3$ as the translational velocity of $Y$ in frame $X$.
Four coordinate frames are defined in Fig. \ref{Notation}, 
including the Body frame $B$ (aligned to the IMU frame), 
the camera frame $C$, the object frame $O$, and the navigation frame $N$.
The frame $N$, also known as east-north-up (ENU), 
is the global reference for the system. 
The GNSS-IMU extrinsic parameter $\boldsymbol{T}_{BG}$ 
and the Camera-IMU extrinsic parameter $\boldsymbol{T}_{BC}$ have been calibrated.
The body pose $\boldsymbol{T}_{NB}$, 
the object pose $\boldsymbol{T}_{NO}$, 
the map point $\boldsymbol{P}_{M}$, 
and the object point $\boldsymbol{P}_{O}$ are the states to be estimated. 
The states also include the body velocity $\boldsymbol{v}_{NB}$ and the velocity $\boldsymbol{v}_{NO}$, the yaw speed $v_{\psi}$, and the scale $s$ of the object.

\subsection{Vision Front-End with Object Processing} 
The object's features are extracted by the method of \cite{shiGoodFeaturesTrack1994a}, 
and the association via features matching uses high-efficiency Lucas-Kanade (LK) optical flow.
However, it is not easy to track the object accurately even using multiscale pyramidal optical flow,  
due to the motion of objects. 
To improve the association robustness, 
we use a method named optical flow with the descriptor.  
Firstly a bidirectional optical flow is used for features matching,  
then the BRIEF descriptors \cite{rubleeORBFeature-2011} are calculated 
to select good matches via descriptor distance. 

If the stationary objects are determined quickly, 
more available static features could be used for SLAM. 
As the multi-view geometry constraint does not satisfy the dynamic features, 
the standard deviation (STD) of the triangulated depth sequence $ \boldsymbol{d}= [d_0,d_1,\cdots,d_m] $ can be used to determine the motion state. 
The STD should be small for static objects, 
while large for dynamic ones. 

\subsection{Prior Height Model for Object Parametrization}

TwistSLAM \cite{TwistSLAM-2022} set the object pose as 3 DoF with plane road constraints, 
which may affect the accuracy on slopes. 
Since most small slopes (such as Fig. \ref{Abstract} shows) could lead to the long-term displacement of the z-axis but a slight change of pitch and roll, 
we define the object pose as 4 DoF, i.e., 3D translation and 1D rotation (yaw).
The initial yaw $\psi$ can be determined with $\psi = tan^{-1} \left( {v_n}/{v_e} \right)$ \cite{jinFastAccurateInitialization2021}, 
where $v_n$ and $v_e$ are the north and east components of the object's initial velocity respectively. 
The initial velocity can be calculated by position differential.
Therefore, the initial position is the key to determining the initial state of the object. 
However, the object position can not be calculated by triangulation due to scale ambiguity.
\begin{figure}[!t]
	\centering
	\includegraphics[width=0.9\columnwidth]{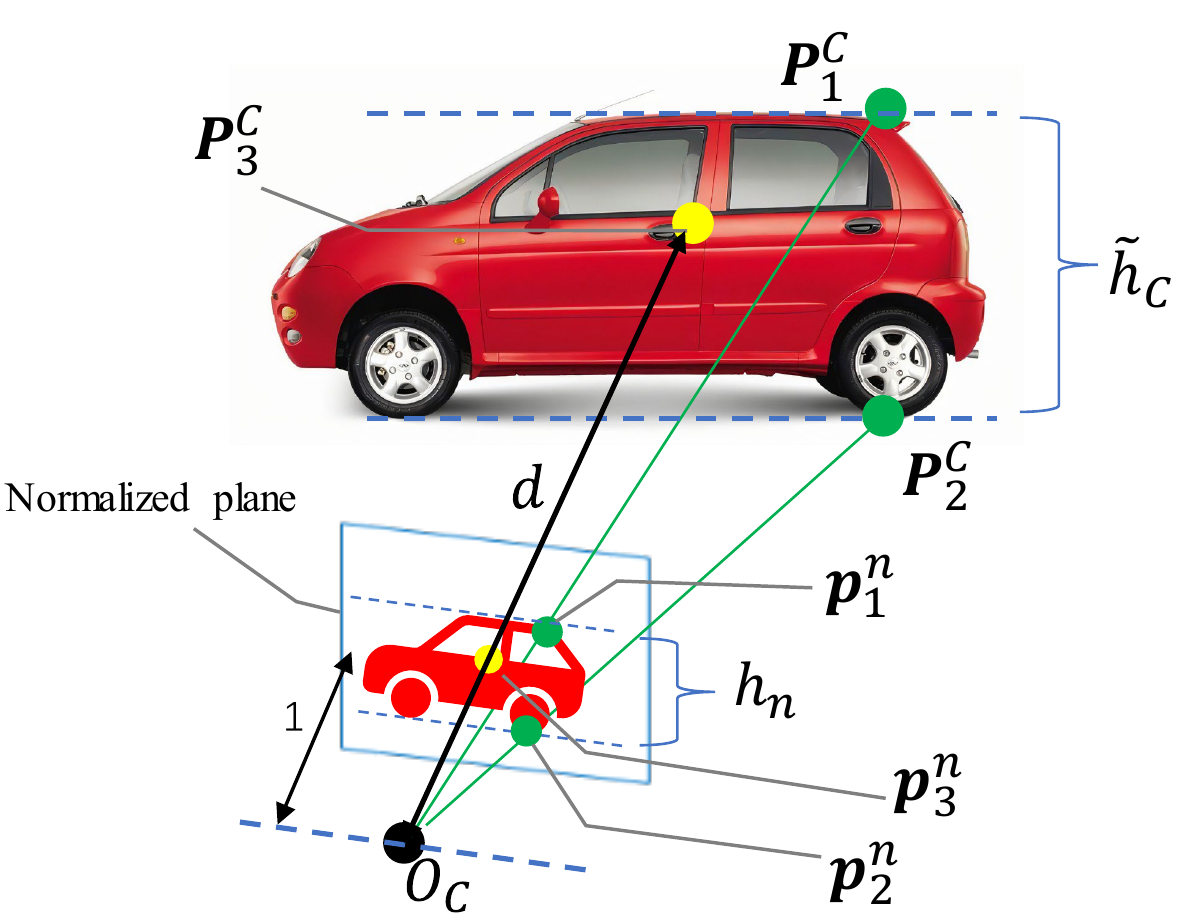}
	\caption{Prior height Model of the object. 
	$\boldsymbol{O}_C$ is the center of the frame $C$, 
	$\boldsymbol{P}_i^C$ is the point of the object in the frame $C$, 
	$\boldsymbol{p}_i^n$ is the projection of $\boldsymbol{P}_i^C$ on the normalized plane.
	The green points $\boldsymbol{P}_1^C$ and $\boldsymbol{P}_2^C$ are the top and bottom points of the object respectively, 
	and the prior height $\tilde{h}_C$ represents the distance of them.
	The yellow point $\boldsymbol{P}_3^C$ is projected to the center of the normalized plane.
	$\boldsymbol{O}_C\boldsymbol{P}_3^C = d$ is object depth in the frame $C$.}
	\label{PriorH}
\end{figure}

To solve this problem, 
we propose a prior height model shown in Fig. \ref{PriorH}.
Assuming the height of a class (such as a person or car) is known, 
the initial position of the object in the frame $C$ can be determined. 
Let $ \boldsymbol{P}_1^C(X_1^C,Y_1^C,Z_1^C) $ and $ \boldsymbol{P}_2^C(X_2^C,Y_2^C,Z_2^C) $ be the top and bottom point of the object in the frame $C$ respectively, 
they are projected as $ \boldsymbol{p}_1^n(x_1^n,y_1^n) $ and $ \boldsymbol{p}_2^n(x_2^n,y_2^n) $ on the normalized plane. 
$ \boldsymbol{p}_i^n(x_i^n,y_i^n) $ can be converted via reprojection from the image observation $ (u_i,v_i) $.
According to the triangular similarity, we have:
\begin{equation}
\begin{aligned}
        \frac{1}{d} = \frac{y_2^n-y_1^n}{Y_2^C-Y_1^C} = \frac{y_2^n-y_1^n}{\tilde{h}_C}
\end{aligned}
\end{equation}
where $\tilde{h}_C $ is the prior height of the object, 
then we have the depth $d = \tilde{h}_C/(y_2^n-y_1^n)$.
Hence, the coordinates of any object point $k$ in the frame $C$ can be calculated:
\begin{equation}
\begin{aligned}
        \left[ \begin{matrix}
                        X_k^C & Y_k^C & Z_k^C
                \end{matrix}\right]  = \frac{\tilde{h}_C}{y_2^n-y_1^n}\cdot
        \left[ \begin{matrix}
                        x_k^n & y_k^n & 1
                \end{matrix}\right]
\end{aligned}
\end{equation}
Assuming the scale $s = \hat{h}_C/\tilde{h}_C$, where $\hat{h}_C$ is the true height.
$ s $ is added to the state vector for further refinement.

\subsection{Factor Graph Optimization for Vision/INS/GNSS integration with Object Tracking}\label{VIGoptimization}
\begin{table*}
	\caption{Camera Pose Comparison with Existing Algorithms on the KITTI Dataset.
		ATE is in $m$, RPE$_t$ in $m/f$, RPE$_R$ in $^{\circ}/f$}
	%%% \tablesize{} %% You can specify the fontsize here, e.g., \tablesize{\footnotesize}. If commented out \small will be used.
	\centering
	\resizebox{2.05\columnwidth}{!}{
		\begin{tabular}{c|ccc|ccc|ccc|ccc}
			\toprule
			\multirow{2}{*}{seq} & \multicolumn{3}{c|}{VDO-SLAM \cite{VDO-SLAM-2021}} & \multicolumn{3}{c|}{DynaSLAM II \cite{DynaSLAMII-2021}} & \multicolumn{3}{c|}{TwistSLAM \cite{TwistSLAM-2022}} & \multicolumn{3}{c}{Ours}                                                                                                                     \\
			                     & ATE                                                & RPE$_t$                                                  & RPE$_R$                                              & ATE                      & RPE$_t$ & RPE$_R$       & ATE & RPE$_t$        & RPE$_R$        & ATE            & RPE$_t$        & RPE$_R$       \\
			\midrule
			00                   & -                                                  & 0.07                                                     & 0.07                                                 & 1.29                     & 0.04    & 0.06          & -   & 0.04           & 0.05           & \textbf{0.03}  & \textbf{0.02}  & \textbf{0.04} \\
			01                   & -                                                  & 0.12                                                     & 0.04                                                 & 2.31                     & 0.05    & 0.04          & -   & 0.04           & \textbf{0.03}  & \textbf{0.05}  & \textbf{0.03}  & 0.05          \\
			02                   & -                                                  & 0.04                                                     & \textbf{0.02}                                        & 0.91                     & 0.04    & \textbf{0.02} & -   & \textbf{0.03}  & 0.03           & \textbf{0.05}  & \textbf{0.03}  & 0.06          \\
			03                   & -                                                  & 0.08                                                     & 0.03                                                 & 0.69                     & 0.06    & 0.04          & -   & 0.06           & \textbf{0.02}  & \textbf{0.02}  & \textbf{0.02}  & 0.05          \\
			04                   & -                                                  & 0.11                                                     & 0.05                                                 & 1.42                     & 0.07    & 0.06          & -   & 0.06           & \textbf{0.04}  & \textbf{0.04}  & \textbf{0.02}  & 0.05          \\
			05                   & -                                                  & 0.09                                                     & \textbf{0.02}                                        & 1.34                     & 0.06    & 0.03          & -   & 0.06           & \textbf{0.02}  & \textbf{0.02}  & \textbf{0.01}  & 0.03          \\
			06                   & -                                                  & 0.02                                                     & 0.05                                                 & 0.19                     & 0.02    & \textbf{0.04} & -   & 0.02           & \textbf{0.04}  & \textbf{0.01}  & \textbf{0.01}  & \textbf{0.04} \\
			07                   & -                                                  & -                                                        & -                                                    & 3.10                     & 0.05    & 0.07          & -   & \textbf{0.04}  & \textbf{0.04}  & \textbf{0.09}  & \textbf{0.04}  & \textbf{0.04} \\
			08                   & -                                                  & -                                                        & -                                                    & 1.68                     & 0.10    & 0.04          & -   & 0.07           & \textbf{0.03}  & \textbf{0.07}  & \textbf{0.05}  & 0.08          \\
			09                   & -                                                  & -                                                        & -                                                    & 5.02                     & 0.06    & 0.06          & -   & 0.05           & \textbf{0.04}  & \textbf{0.02}  & \textbf{0.01}  & \textbf{0.04} \\
			10                   & -                                                  & -                                                        & -                                                    & 1.30                     & 0.07    & \textbf{0.03} & -   & 0.07           & \textbf{0.03}  & \textbf{0.03}  & \textbf{0.02}  & 0.04          \\
			11                   & -                                                  & -                                                        & -                                                    & 1.03                     & 0.04    & 0.03          & -   & 0.03           & \textbf{0.02}  & \textbf{0.08}  & \textbf{0.02}  & 0.04          \\
			13                   & -                                                  & -                                                        & -                                                    & 1.10                     & 0.04    & 0.04          & -   & \textbf{0.03}  & \textbf{0.04}  & \textbf{0.17}  & 0.07           & 0.08          \\
			14                   & -                                                  & -                                                        & -                                                    & 0.12                     & 0.03    & 0.08          & -   & 0.03           & \textbf{0.06}  & \textbf{0.03}  & \textbf{0.02}  & 0.07          \\
			18                   & -                                                  & 0.07                                                     & 0.02                                                 & 1.09                     & 0.05    & 0.02          & -   & \textbf{0.04}  & \textbf{0.02}  & \textbf{0.05}  & 0.05           & 0.03          \\
			19                   & -                                                  & -                                                        & -                                                    & 2.25                     & 0.05    & 0.03          & -   & 0.03           & 0.03           & \textbf{0.29}  & \textbf{0.02}  & \textbf{0.01} \\
			20                   & -                                                  & 0.17                                                     & \textbf{0.03}                                        & 1.36                     & 0.07    & 0.04          & -   & \textbf{0.04}  & \textbf{0.03}  & \textbf{0.07}  & \textbf{0.04}  & 0.06          \\
			\midrule
			mean                 & -                                                  & 0.087                                                    & 0.037                                                & 1.541                    & 0.053   & 0.043         & -   & 0.044          & \textbf{0.034} & \textbf{0.066} & \textbf{0.028} & 0.048         \\
			std                  & -                                                  & 0.044                                                    & 0.018                                                & 1.159                    & 0.019   & 0.017         & -   & \textbf{0.015} & \textbf{0.011} & \textbf{0.068} & 0.016          & 0.018         \\
			\bottomrule
		\end{tabular}
	}
	\label{CamPoseTable}
\end{table*}

As mentioned in section \ref{NotationSect}, 
the states can be defined as 
$\boldsymbol{X} = [\boldsymbol{T}_{NB},\boldsymbol{v}_{NB},\boldsymbol{P}_{M}, 
s,\boldsymbol{T}_{NO},\boldsymbol{P}_{O},v_{\psi},\boldsymbol{v}_{NO}]$, 
Assuming all the observations are with Gaussian distribution, 
the factors can be processed with one optimizer.
The loss function of DynaVIG is defined as follows:
\begin{equation}\label{DynaVIGOpt}
	\begin{aligned}
		\mathop{\arg\min}_{\boldsymbol{X}} = 
		\biggl\{ &
		\sum{\left\Vert\boldsymbol{e}_{sta}\right\Vert^2_{\Sigma_S}} +
		\sum{\left\Vert \boldsymbol{e}_{IMU}\right\Vert^2_{\Sigma_I}} + \\ &
		\sum{\left\Vert \boldsymbol{e}_{GNSS} \right\Vert^2_{\Sigma_G}}+
		\sum{\left\Vert \boldsymbol{e}_{obj} \right\Vert^2_{\Sigma_O}}+ \\ &
		\sum{\left\Vert \boldsymbol{e}_{dm} \right\Vert^2_{\Sigma_{\boldsymbol{v}}}}
		\biggr\}
	\end{aligned}
\end{equation}
where $\boldsymbol{e}_{sta},\boldsymbol{e}_{IMU},\boldsymbol{e}_{GNSS} $,  $\boldsymbol{e}_{obj}$, and $\boldsymbol{e}_{dm}$ are the static feature, 
IMU pre-integration,  GNSS, object feature, and object dynamics factors respectively, 
$\Sigma $ is the covariance matrix.
The traditional $\boldsymbol{e}_{sta},\boldsymbol{e}_{IMU},\boldsymbol{e}_{GNSS} $ are as follows:
\begin{equation}\label{VIG_factors}
	\begin{aligned}
		\boldsymbol{e}_{sta} &=  \pi((\boldsymbol{T}_{NB}\boldsymbol{T}_{BC})^{-1}\boldsymbol{P}_{M})-\boldsymbol{p}_s \\
		\boldsymbol{e}_{IMU} &=   f_{PI}(\Delta\boldsymbol{R}_{ij},\Delta\boldsymbol{v}_{ij},\Delta\boldsymbol{t}_{ij},\boldsymbol{T}_{NB},\boldsymbol{v}_{NB})  \\
		\boldsymbol{e}_{GNSS}   &=      (\boldsymbol{T}_{NB}\boldsymbol{T}_{BG})|_{\boldsymbol{t}} -  \boldsymbol{t}^N_G
	\end{aligned}
\end{equation}
where $\pi$ is the reprojection function, 
$\boldsymbol{p}_s$ is the coordinate of the static feature on the normalized plane;
$f_{PI}$ is the IMU factor function, 
$\Delta\boldsymbol{R}_{ij},\Delta\boldsymbol{v}_{ij},\Delta\boldsymbol{t}_{ij}$ are the changes of rotation, velocity, and position pre-integrated using IMU measurements respectively;
$\boldsymbol{t}^N_G$ indicates the GNSS position solutions in the frame $N$.

Based on the reprojection $\pi$, 
$\boldsymbol{e}_{obj}$ can be reconstructed:
\begin{equation}
	\begin{aligned}
		\boldsymbol{e}_{obj}=\pi(s\cdot\boldsymbol{T}_{CN}\boldsymbol{T}_{NO}\boldsymbol{P}_{O})-\boldsymbol{p}_O
	\end{aligned}
\end{equation}
where $\boldsymbol{T}_{CN} $ is the inverse of the camera pose; 
$\boldsymbol{P}_{O} $ is the coordinate of the object point, 
$\boldsymbol{p}_O $ is its image observation.

For the object's dynamics model, 
we assume the velocities of the yaw and translation vary slowly. 
Let $\boldsymbol{v}=[v_{\psi},\boldsymbol{v}_{NO}]$, 
it can be modeled as a random constant in a short time:
\begin{equation}\label{RandConstVelCont}
	\begin{aligned}
		\dot{\boldsymbol{v}} &= \boldsymbol{0} \\
		\boldsymbol{v}_{i-1} &= \boldsymbol{v}_i
	\end{aligned}
\end{equation}
The two equations are the continuous and discrete forms of the random process of $\boldsymbol{v}$ respectively, 
$i-1$ and $i$ are two consecutive images.
We propose to set the variance intensity of $\boldsymbol{v}$ according to the motion complexity. 
Assuming the current velocity is the same as the previous, 
the variance intensity could be determined more accurately. 
Therefore, the random model and the optimization are cause and effect to each other.
Then $\boldsymbol{e}_{dm}$ is defined as follows: 
\begin{equation}
	\begin{aligned}
		\boldsymbol{e}_{dm} &= \boldsymbol{v}_i - \boldsymbol{v}_{i-1} \\
		\Sigma_{\boldsymbol{e}_{dm}} &= exp(\left \| \boldsymbol{v}_{i-1} \right \| \cdot K_O)
	\end{aligned}
\end{equation}
where $K_O$ is the gain factor for velocity amplification.

\section{EXPERIMENTAL RESULTS}

\subsection{Experimental Setup}

The proposed DynaVIG is evaluated on the KITTI Tracking dataset \cite{geigerVisionMeetsRobotics2013},  
which mainly contains cars and pedestrians, 
the ground truth of the camera and objects are provided. 
The IMU is extracted from the KITTI Raw dataset since only the raw IMU with a high rate (100Hz) is useful.
The KITTI IMU may contain some sick ranges, 
including time stamp errors and duplicate records. 
Fortunately, these ranges are usually very short (within 0.1s), 
allowing us to fix them by interpolating their neighbors. 
The GNSS measurements are simulated by corrupting the trajectory with Gaussian noise, 
the sampling rate is 1Hz. 
The number of keyframes in the sliding window is limited to 10. 
The factor graph-based GTSAM \cite{FrankFactorGraphs2021} is used for optimization. 

The major evaluation metrics are the absolute trajectory error (ATE) and relative pose error (RPE) \cite{sturm12iros}. 
The object scale error and the computational time are analyzed as well. 
We compare our results with state-of-the-art algorithms.

\subsection{Camera Pose Estimation}

This section analyzes the camera pose estimation.
Table \ref{CamPoseTable} shows the comparison of our method with the existing algorithms. 
The previous works all used stereo vision, 
and only DynaSLAM II calculated the ATE. 
We can see that DynaVIG achieves obvious improvements on the ATE and RPE$_t$, 
indicating that GNSS and INS have good effects on the camera translation. 
The RPE$_R$ of TwistSLAM slightly outperforms DynaVIG. 
Since the KITTI sequences are usually too short for IMU bias to converge, 
the camera rotation depends mainly on the visual observations, 
thus DynaVIG using a monocular camera obtains lower rotation accuracy than stereo systems. 
As discussed in \cite{sturm12iros}, 
the low precision of rotation will further deteriorate RPE$_t$, 
because RPE$_t$ considers both translational and rotational errors. 
Therefore, the higher RPE$_t$ accuracy of DynaVIG shows that multi-sensor fusion has a great advantage in translation estimation,
compared with pure visual SLAM. 

\subsection{Object Tracking}

\begin{table*}
        \caption{Object Pose Comparison with Existing Algorithms on the KITTI Dataset.
                ATE is in $m$, RPE$_t$ in $m/m$, RPE$_R$ in $^{\circ}/m$}
        %%% \tablesize{} %% You can specify the fontsize here, e.g., \tablesize{\footnotesize}. If commented out \small will be used.
        \centering
        \resizebox{2.05\columnwidth}{!}{
                \begin{tabular}{cc|ccc|ccc|ccc|c}
                        \toprule
                        \multirow{2}{*}{seq}      & \multirow{2}{*}{id} & \multicolumn{3}{c|}{DynaSLAM II \cite{DynaSLAMII-2021}} & \multicolumn{3}{c|}{TwistSLAM \cite{TwistSLAM-2022}} & \multicolumn{4}{c}{Ours}                                                                                                                   \\
                                                  &                     & ATE                                                     & RPE$_t$                                              & RPE$_R$                  & ATE            & RPE$_t$       & RPE$_R$        & ATE            & RPE$_t$        & RPE$_R$       & scale error \\
                        \midrule
                        03                        & 1                   & 0.69                                                    & 0.34                                                 & 1.84                     & 0.31           & \textbf{0.10} & \textbf{0.28}  & \textbf{0.26}  & 0.73           & 0.47          & 17.38\%     \\
                        \midrule
                        05                        & 31                  & 0.51                                                    & 0.26                                                 & 13.50                    & 0.35           & \textbf{0.19} & 0.58           & \textbf{0.18}  & 0.24           & \textbf{0.32} & 3.66\%      \\
                        \midrule
                        10                        & 0                   & 0.95                                                    & 0.40                                                 & 2.84                     & 0.77           & 0.21          & 1.98           & \textbf{0.12}  & \textbf{0.20}  & \textbf{0.43} & 13.17\%     \\
                        \midrule
                        \multirow{2}{*}{11}       & 0                   & 1.05                                                    & 0.43                                                 & 12.51                    & 0.17           & \textbf{0.23} & \textbf{0.23}  & \textbf{0.16}  & 0.74           & 0.46          & 5.06\%      \\
                                                  & 35                  & 1.25                                                    & 0.89                                                 & 16.64                    & 0.10           & 0.03          & \textbf{0.11}  & \textbf{0.07}  & \textbf{0.02}  & 0.27          & -           \\
                        \midrule
                        \multirow{2}{*}{18}       & 2                   & 1.10                                                    & 0.30                                                 & 9.27                     & 0.21           & \textbf{0.27} & 0.66           & \textbf{0.05}  & 0.32           & \textbf{0.29} & 1.14\%      \\
                                                  & 3                   & 1.13                                                    & 0.55                                                 & 20.05                    & \textbf{0.15}  & 0.21          & 0.56           & 0.26           & \textbf{0.17}  & \textbf{0.22} & 10.22\%     \\
                        \midrule
                        \multirow{2}{*}{19}       & 63                  & 0.86                                                    & 1.45                                                 & 48.80                    & \textbf{0.28}  & 2.17          & 1.08           & 0.44           & \textbf{0.24}  & \textbf{0.25} & -           \\
                                                  & 72                  & 0.99                                                    & 1.12                                                 & 3.36                     & 0.16           & 0.05          & 0.34           & \textbf{0.11}  & \textbf{0.01}  & \textbf{0.08} & 93.10\%     \\
                        \midrule
                        \multirow{3}{*}{20}       & 0                   & 0.56                                                    & 0.45                                                 & 1.30                     & \textbf{0.17}  & \textbf{0.20} & 0.72           & 0.23           & 0.69           & \textbf{0.28} & 7.36\%      \\
                                                  & 12                  & 1.18                                                    & 0.40                                                 & 6.19                     & 0.24           & \textbf{0.20} & 1.54           & \textbf{0.06}  & 0.36           & \textbf{0.60} & 10.40\%     \\
                                                  & 122                 & 0.87                                                    & 0.72                                                 & 5.75                     & 0.17           & \textbf{0.02} & \textbf{0.07}  & \textbf{0.11}  & 0.48           & 0.55          & 11.11\%     \\
                        \midrule
                        \multicolumn{2}{c|}{mean} & 0.928               & 0.609                                                   & 11.838                                               & 0.257                    & \textbf{0.323} & 0.679         & \textbf{0.170} & 0.350          & \textbf{0.352} & 17.26\%                     \\
                        \multicolumn{2}{c|}{std}  & 0.240               & 0.369                                                   & 13.140                                               & 0.177                    & 0.588          & 0.587         & \textbf{0.114} & \textbf{0.258} & \textbf{0.151} & 27.07\%                     \\
                        \bottomrule
                \end{tabular}
        }
        \label{ObjPoseTable}
\end{table*}

\begin{figure*}[!t]
        \centering
        \includegraphics[width=0.99\textwidth]{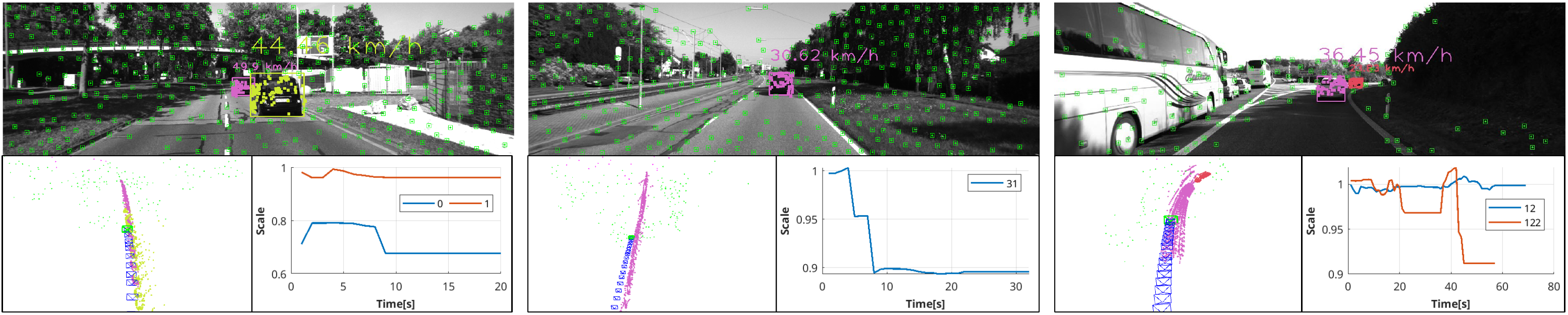}
        \caption{Visualization of object tracking of the sequence 0003 (left), 0005 (Middle), and 0020 (Right) on the KITTI trakcing dataset.
                (Top): Pink and yellow rectangles are objects with speed. Green points with rectangles are environment features; 
                (Bottom left): 4D point clouds of the objects and map; 
                (Bottom right): object scale converging curves with time.}
        \label{Map_Scale}
\end{figure*}

For object pose,  
we use the sequences analyzed in DynaSLAM II and TwistSLAM, 
the results are shown in Table \ref{ObjPoseTable}.
It demonstrates that DynaVIG mostly outperforms other algorithms on the ATE. 
This also benefits from the advantage of multi-sensor fusion in translation estimation, 
although the object translation of DynaVIG contains a scale error. 
This is because the estimation of object pose largely depends on the camera pose when optimized jointly, 
as the camera pose estimation has more sensors and a better geometry structure of features. 
DynaVIG achieves better RPE$_R$, 
which is greatly due to the proposed dynamics model, 
good examples are objects with speed or/and direction changes, 
such as 10-0, 20-0, and 20-12 (sequence-id).
Compared with TwistSLAM, 
the slightly worse RPE$_t$ of DynaVIG is most likely due to the scale error.

The monocular scale errors of the objects are calculated in Table \ref{ObjPoseTable}, 
which is about 10\% in most cases.
The scale converging curves are shown at the bottom right of each column in Fig. \ref{Map_Scale}, 
the scale of most objects can converge with time. 
However, there are some abnormal cases.
Both 11-35 and 19-63 are static cars, 
and the camera of both sequences stayed stationary for a long time during the driving. 
The lack of translation leads to the inability to effectively scale estimation.
20-122 takes a small size of the image and is blocked for a short time, 
thus its scale may not converge. 

\subsection{Computational Time}
% \begin{table}[!t]
% 	\caption{Computational Efficiency Compared with DynaSLAM II.}
% 	\centering
% 	\renewcommand{\arraystretch}{1.5}
% 	\begin{tabular}{c|c|c|c}
% 		\toprule
% 		\multicolumn{2}{c|}{ } & DynaSLAM II & Ours           \\
% 		\midrule
% 		\multirow{2}{*}{3}     & front-end   & 80.10 & 84.85  \\
% 		                       & back-end    & 61.37 & 148.83 \\
% 		\midrule
% 		\multirow{2}{*}{20}    & front-end   & 94.56 & 86.89  \\
% 		                       & back-end    & 65.03 & 166.35 \\
% 		\bottomrule
% 	\end{tabular}
% 	\label{CompTime}
% \end{table}

\begin{table}[!t]
	\caption{Comparison of Computational Time ($mSec$).}
	\centering
	\renewcommand{\arraystretch}{1.5}
	\begin{tabular}{c|cc|cc}
		\toprule
		            & \multicolumn{2}{c|}{3} & \multicolumn{2}{c}{20}                                   \\
		            & front-end              & back-end               & front-end      & back-end       \\
		\midrule
		DynaSLAM II & \textbf{80.10}         & \textbf{61.37}         & 94.56          & \textbf{65.03} \\
		\midrule
		Ours        & 84.85                  & 148.83                 & \textbf{86.89} & 166.35         \\
		\bottomrule
	\end{tabular}
	\label{CompTime}
\end{table}

In this section, we evaluate the computational time of our method.
The experiments are carried out on a desktop PC with an Intel i3-4150 at 3.5GHz and 16-GB memory.
To perform a fair comparison with DynaSLAM II, 
the front-end including the object processing of DynaVIG is treated as the tracking thread in DynaSLAM II, 
and the back-end is treated as the Local BA thread of DynaSLAM II. 
The results are listed in Table \ref{CompTime}. 
The front-end of DynaVIG spends about the same amount of time as DynaSLAM II, 
showing that both DynaVIG and DynaSLAM II could run in real time. 
However, the back-end of DynaVIG costs more much time than DynaSLAM II. 
Due to the different number of cameras and features, 
the influences on the front-end computation are hard to compare. 
As for the back-end, 
the number of features should be likely the main reason for the difference in computational time.

\section{CONCLUSIONS}

We propose DynaVIG, 
a navigation and object tracking system based on the Monocular Vision/INS/GNSS integration,
which can eliminate the drift of traditional SLAM and realize 3D object tracking with a monocular camera. 
A prior height model is proposed for pose initialization and scale estimation of the object, 
and an accurate dynamics model is constructed for precise tracking of objects with complex motion.
Compared with the existing algorithms, 
DynaVIG achieves high-precision navigation and object tracking with real-time performance.
In summary, DynaVIG is one of the state-of-the-art research of dynamic SLAM with object tracking.
To the best of our knowledge, 
this is the first study using multi-sensor integration for accurate global navigation and object tracking.

% \addtolength{\textheight}{-10cm}   % This command serves to balance the column lengths

%%%%%%%%%%%%%%%%%%%%%%%%%%%%%%%%%%%%%%%%%%%%%%%%%%%%%%%%%%%%%%%%%%%%%%%%%%%%%%%%

%%%%%%%%%%%%%%%%%%%%%%%%%%%%%%%%%%%%%%%%%%%%%%%%%%%%%%%%%%%%%%%%%%%%%%%%%%%%%%%%



\section*{ACKNOWLEDGMENT}

Thanks to Tianyi Liu and Shaoquan Feng from Wuhan University, 
for their valuable suggestions of algorithm improvements and data processing.

%%%%%%%%%%%%%%%%%%%%%%%%%%%%%%%%%%%%%%%%%%%%%%%%%%%%%%%%%%%%%%%%%%%%%%%%%%%%%%%%

\end{document}